\documentclass{article} 
\usepackage{bm}
\usepackage{nips14submit_e,times}
\usepackage{url}
\usepackage{amssymb}
\usepackage{amsthm}
\usepackage[numbers]{natbib}
\usepackage{enumitem}
\usepackage{amsmath}  
\usepackage{array}
\usepackage{multirow}

\usepackage{algorithm}
\usepackage{algorithmic}

\usepackage{graphicx}

\title{Auxiliary Multimodal LSTM for Audio-visual Speech Recognition and Lipreading}

\author{
Chunlin Tian,
Weijun Ji\\
University of the Chinese Academy of Sciences,19A Yuquanlu, Beijing, 100049, P.R.China\\
{\tt\small \{tianchunlin15, jiweijun15\}@mails.ucas.ac.cn}
}

\nipsfinalcopy 

\begin{document}

\maketitle

\begin{abstract}
The \emph{Aduio-visual Speech Recognition} (AVSR) which employs both the video and audio information to do \emph{Automatic Speech Recognition} (ASR) is one of the application of multimodal leaning making ASR system more robust and accuracy. The traditional models usually treated AVSR as inference or projection but strict prior limits its ability. As the revival of deep learning, \emph{Deep Neural Networks} (DNN) becomes an important toolkit in many traditional classification tasks including ASR, image classification, natural language processing. Some DNN models were used in AVSR like \emph{Multimodal Deep Autoencoders} (MDAEs), \emph{Multimodal Deep Belief Network} (MDBN) and \emph{Multimodal Deep Boltzmann Machine} (MDBM) that actually work better than traditional methods. However, such DNN models have several shortcomings: (1) They don't balance the modal fusion and temporal fusion, or even haven't temporal fusion; (2)The architecture of these models isn't end-to-end, the training and testing getting cumbersome. We propose a DNN model, \emph{Auxiliary Multimodal LSTM} (am-LSTM),  to overcome such weakness. The am-LSTM could be trained and tested in one time, alternatively easy to train and preventing overfitting automatically. The extensibility and flexibility are also take into consideration. The experiments shows that am-LSTM is much better than traditional methods and other DNN models in three datasets: AVLetters, AVLetters2, AVDigits.
\end{abstract}

\section{Introduction}

\label{intro}

\emph{Automatic Speech Recognition} (ASR) has been investigated over several years, and there is a wealth of literature. The recent progress is \emph{Deep Speech2} \cite{amodei2015deep}, which utilizes deep \emph{Convolution Neural Network} (CNN)\cite{krizhevsky2012imagenet}, LSTM\cite{hochreiter1997long} and CTC \cite{graves2006connectionist}, and sequence-to-sequence models \cite{sutskever2014sequence}. Although ASR achieved excellent result, it had some intrinsical problems including insufficient tolerance of noise and disturbance. Besides, some illusion occurs when the auditory component of one sound is paired with the visual component of another sound, leading to the perception of a third sound \cite{nath2012a}. This was described as McGurk effect (McGurk \& MacDonald 1976).

Benefited from the multimodal learning, \emph{Audio-visual Speech Recognition} (AVSR) is a supplement to ASR that mixes audio and visual information together, and amount of work verified that AVSR strengthened the ASR system \cite{ngiam2011multimodal} \cite{amer2014multimodal}. When it comes to AVSR, the core part is multimodal learning. In the early decades, many models aimed to fuse the multimodal data more representative and discriminative, which contain multimodal extensions of \emph{Hidden Markov Models} (HMMs) and some statistical models. But strong prior assumption limits such models. As the revival of \emph{Neural Networks}, some deep models were proposed in multimodal learning. Different from traditional models, deep models usually have two main aspects:

(1) Many researchers think of \emph{Deep Neural Networks} (DNN) as performing a kind of good representation learning \cite{bengio2015deep}, that is, the ability to extract feature of DNN can be exploited for varieties of tasks, especially the CNN for image feature extraction. For aural information, it is refined by some deep models including \emph{Deep Belief Network} (DBN), \emph{Deep Autoencoder} (DAE), \emph{Restricted Boltzmann Machines} (RBMs) and \emph{Deep Bottleneck Features} (DBNF)\cite{mroueh2015deep}\cite{noda2015audio-visual}\cite{tamura2015audio}.

(2) The other merit of deep models is the well-performed fusion, conquering the biggest disadvantage of traditional methods. \emph{Multimodal DBN} (MDBN)\cite{srivastava2012learning}, \emph{Multimodal DAEs} (MDAEs)\cite{ngiam2011multimodal}, \emph{Multimodal Deep Boltzmann Machine} (MDBM)\cite{srivastava2012multimodal} are all unsupervised DNN that perform fusion.

However, the aforementioned deep models have two primary shortcomings:

(1) They don't balance the modal fusion and temporal fusion, or even haven't temporal fusion. Many methods simply concatenate the features of frames in a single video to do temporal fusion. Besides, in the modal fusion, such methods don't consider the correlation among different frames.

(2) The architecture of these models isn't end-to-end, the training and testing getting cumbersome.

Inspiring by \emph{Multimodal Recurrent Neural Networks} \cite{mao2014deep} for image captioning, we propose an end-to-end DNN architecture, \emph{Auxiliary Multimodal LSTM} (am-LSTM), for AVSR and lipreading. The am-LSTM overcomes the two main weaknesses mentioned before. It is composed of LSTMs, projection and recognition, and is trained once. Therefore the modal fusion and temporal fusion are accomplished at the same time, i.e. the modal and temporal fusion are mixed and combined in terms of balance of the two fusion processes. To avoid overfitting and make the DNN converge faster, we use the connection strategy like \emph{Deep Residual Learning} \cite{he2015deep}--auxiliary connection. Early stopping \cite{prechelt1998automatic} and dropout \cite{srivastava2014dropout} regularization are used to fight against overfitting as well. Because of the CNN and LSTM, am-LSTM is also a type of well-known \emph{CNN-RNN architecture}.

We conducted the experiments in three AVSR datasets: AVLetters (Patterson et al., 2002), AVLetters2 (Cox et al., 2008) and AVDigits (Di Hu et al., 2015). The results suggest that am-LSTM is better than classic AVSR models and some deep models before.

\section{Related work}
\label{sec:related}
\textbf{Traditional AVSR Systems.} Humans understand the multimodal world in a seemingly effortless manner, although there are vast information processing resources dedicated to the corresponding tasks by the brain \cite{maragos2008multimodal}. Whereas it is difficult for computers to understand multimodal information, because different modalities have their character in nature, same information can be expressed in different modalities in very dissimilar way, let alone distinct information. Audio is one dimensional temporal signal, video is three dimensional temporal signal and text carries semantic information.

Multimodal fusion is the most important part of multimodal learning. There exists three kinds of fusion strategies: early fusion, late fusion and hybrid fusion. For early fusion, it suffices to concatenate all monomodal features into a single aggregate multimodal descriptor. In late fusion, each modality is classified independently. Intergration is done at the decision level and is usually based on heuristic rules. Hybrid fusion lies in-between early and late fusion methods and are specifically geared towards modeling multimodal time-evolving data \cite{maragos2008multimodal} \cite{atrey2010multimodal}.

The representative work in the early years in this field is \emph{multistream HMMs} (mHMMs) \cite{nefian2002dynamic} which were confirmed adaptability and flexibility for modeling sequence and temporal data \cite{rabiner1989tutorial}. But probabilistic models have some explicit limitations, especially strong priori. Researchers are also interested in the manner of mapping data in dissimilar space to one space jointly.

\textbf{Deep Learning for AVSR.} Deep learning provides a powerful toolkit for machine learning. In AVSR, DNN also displays obvious increasement of accuracy. \cite{noda2015audio-visual} uses pre-trained CNN to extracted visual features and denoising autoencoders to improve aural features, and then mHMMs to do fusion and classfication. As I mentioned in \ref{intro}, it mainly utilizes the feature extraction of DNN.

Some other methods are multimodal fusion based on unsupervised DNN. One type is based on DAEs, the representative work is MDAEs \cite{ngiam2011multimodal}. Others based on \emph{Boltzmann Machines}, \cite{srivastava2012learning} uses MDBN, \cite{srivastava2012multimodal} uses MDBM. Generally, DAEs-based models are easy to trained, but lack of theoretical support and flexibility. RBM-based models are hard to train because of the partial function \cite{bengio2015deep}, but sufficient support from probabilistic models and simple to extend.

\section{The Proposed Model}

The am-LSTM aims at fusing the audio-visual data at the same time, considering the modal fusion, temporal fusion and the connection between frames simultaneously. In this section, we will introduce the am-LSTM, and show its simplicity and extensity.

\subsection{Simple LSTM}
There are two widely known issues with properly training vanilla RNN, the vanishing and the exploding gradient \cite{pascanu2013difficulty}. Without some training tricks in training RNN, LSTM uses gates to avoid gradient vanishing and exploding. A typical LSTM often has 3 gates: input gate, output gate and forget gate which slows down the disappearance of past information and makes \emph{Backpropagation through Time} (BPTT) easier. The gates work as follows:

\begin{equation}
i_t = sigmoid(W^{i}x_t+U^{i}h_{t-1}+b^i)
\end{equation}
\begin{equation}
f_t = sigmoid(W^{f}x_t+U^{f}h_{t-1}+b^f)
\end{equation}
\begin{equation}
z_t = tanh(W^{z}x_t+U^{z}h_{t-1}+b^z)
\end{equation}
\begin{equation}
c_t = z_{t}i_t+c_{t-1}f_t
\end{equation}
\begin{equation}
o_t = sigmoid(W^{o}x_t+U^{o}h_{t-1}+b^o)
\end{equation}
\begin{equation}
h_t = c_{t}tanh(o_t)
\end{equation}

\begin{figure}[htbp]
\begin{center}
\includegraphics[width=0.8\linewidth]{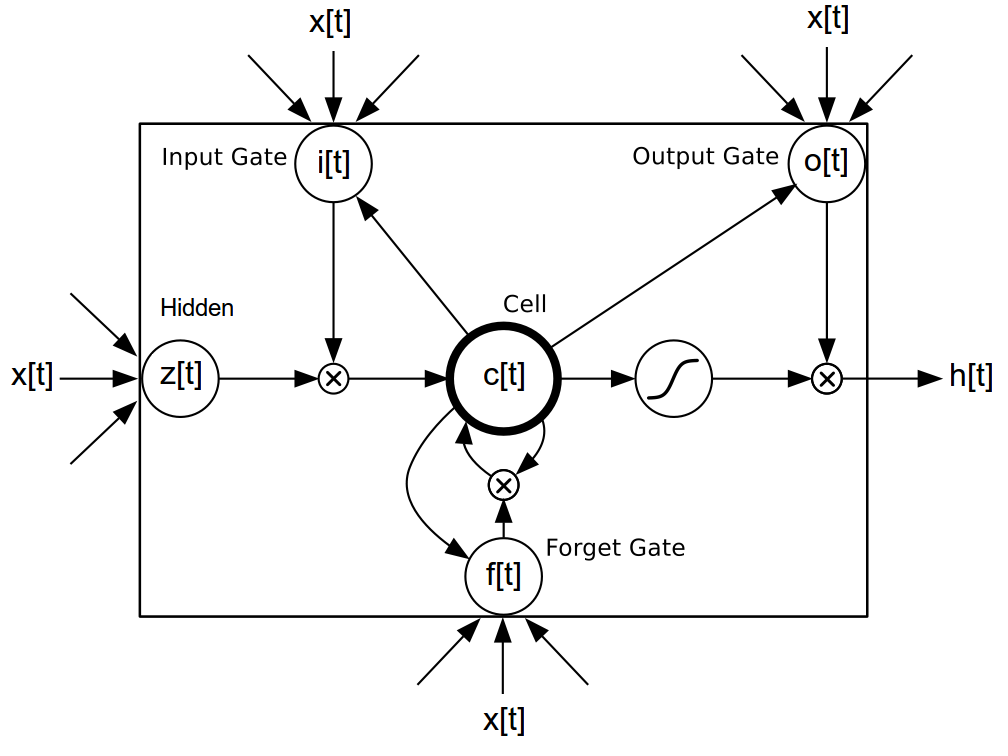}\\
\end{center}
\caption{Simple LSTM}
\end{figure}

\subsection{am-LSTM}

\begin{figure}[htbp]
\begin{center}
\includegraphics[width=1\linewidth]{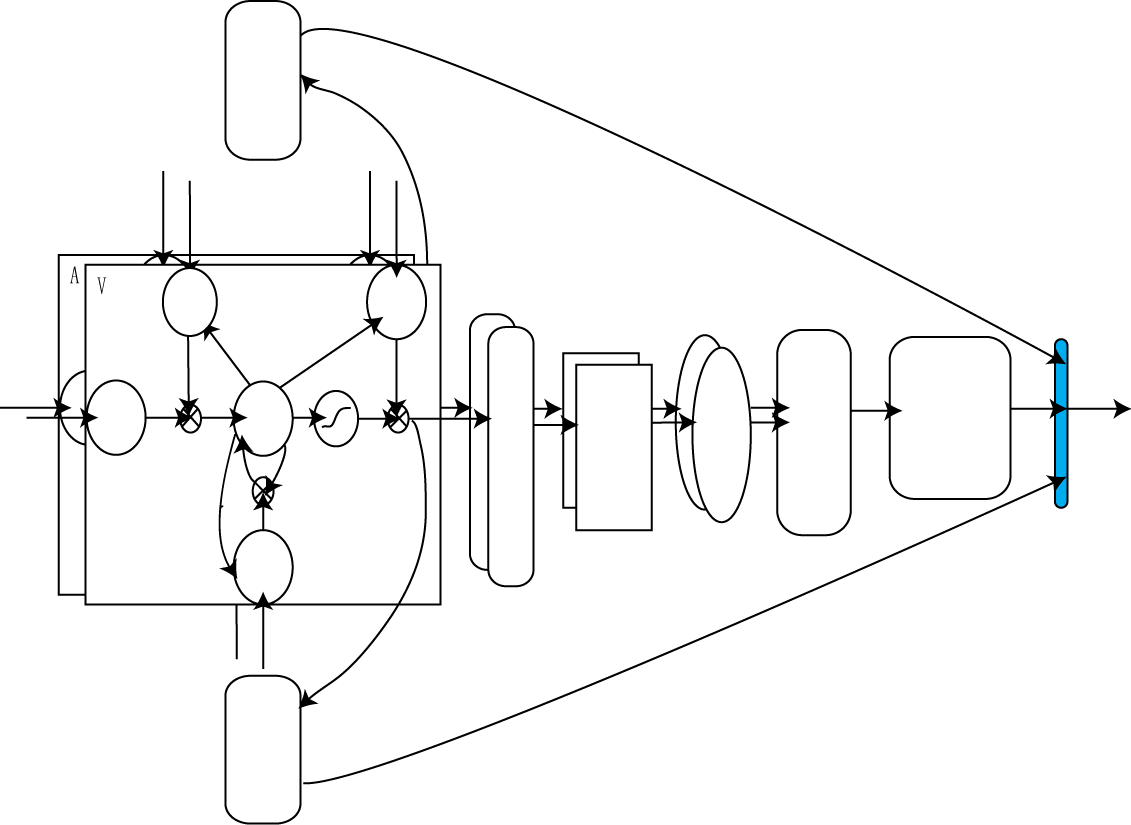}\\
\end{center}
\caption{am-LSTM}
\end{figure}

The am-LSTM is a extension of LSTM which contains two LSTMs and some components. The two LSTMs are video LSTM and audio LSTM, the fundamental architecture is same as simple LSTM, the formulae of video LSTM is (7)$\sim$(12) (the similar formulae can be drawn for tha audio LSTM).
\begin{equation}
i^v_t = sigmoid(W_v^{i}v_t+U_v^{i}h^v_{t-1}+b_v^i)
\end{equation}
\begin{equation}
f^v_t = sigmoid(W_v^{f}v_t+U_v^{f}h^v_{t-1}+b_v^f)
\end{equation}
\begin{equation}
z^v_t = tanh(W_v^{z}v_t+U_v^{z}h^v_{t-1}+b_v^z)
\end{equation}
\begin{equation}
c^v_t = z^v_{t}i^v_t+c^v_{t-1}f^v_t
\end{equation}
\begin{equation}
o^v_t = sigmoid(W_v^{o}v_t+U_v^{o}h^v_{t-1}+b_v^o)
\end{equation}
\begin{equation}
h^v_t = c^v_{t}tanh(o^v_t)
\end{equation}

After video LSTM and audio LSTM, data will be projected into same dimensional space by a projection layer and a activation function thereby. This will also make the model more nonlinear.
\begin{equation}
f_t=g(P^v_tv_t+P^a_ta_t)
\end{equation}

$W_1$ and $W_2$ are projection matrix trained in the DNN, g is the activation function.

\begin{equation}
g(x)=tanh(\frac{2}{3}x)
\end{equation}

The features through video LSTM, audio LSTM and projection could be regarded as well-fused features. The influence of different frames in a single video is summed in both video and audio modality. Then a classical \emph{Multi-layer Perceptron} (MLP) with batch-normalization is used as a classifier.  The training loss is squared multi-label margin loss.

\begin{figure}[htbp]
\begin{center}
\includegraphics[width=1\linewidth]{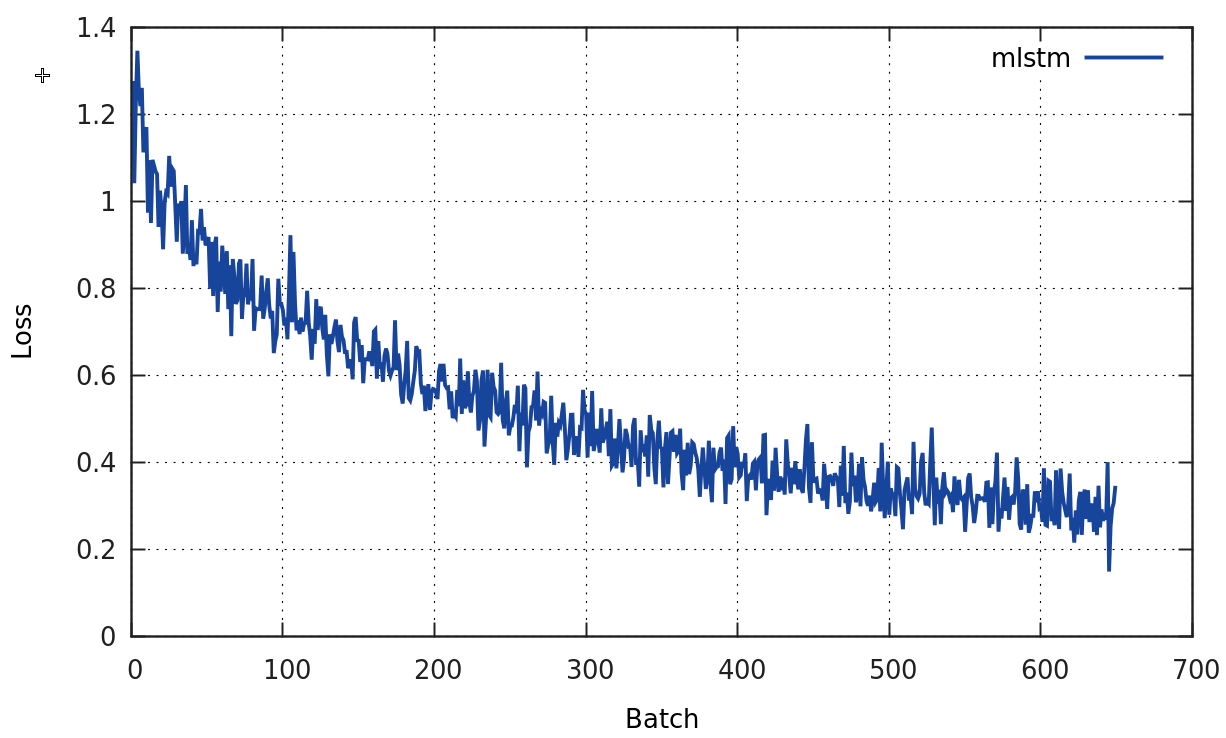}\\
\end{center}
\caption{Batch and loss. We can see that loss is stopped in about 0.2 since strong overfitting.}
\end{figure}
However, the experiment showed that overfitting is very strong in the architecture aforementioned. Hence, we introduce auxiliary connection which accelerates the convergence and prevents overfitting. The auxiliary connection lies after video LSTM and audio LSTM, then data summed and mapped to the target space. In the training, there are three parts taken into account: the main loss, the video auxiliary loss and the audio auxiliary loss.

The implication here is minimizing the audio-visual loss, video loss and audio loss together. The auxiliary networks help the main networks achieve a win-win situation. Like ResNet, the auxiliary networks convey information from original networks, which make am-LSTM consider rich and hierarchy information.

\begin{figure}[htbp]
\begin{center}
\includegraphics[width=0.7\linewidth]{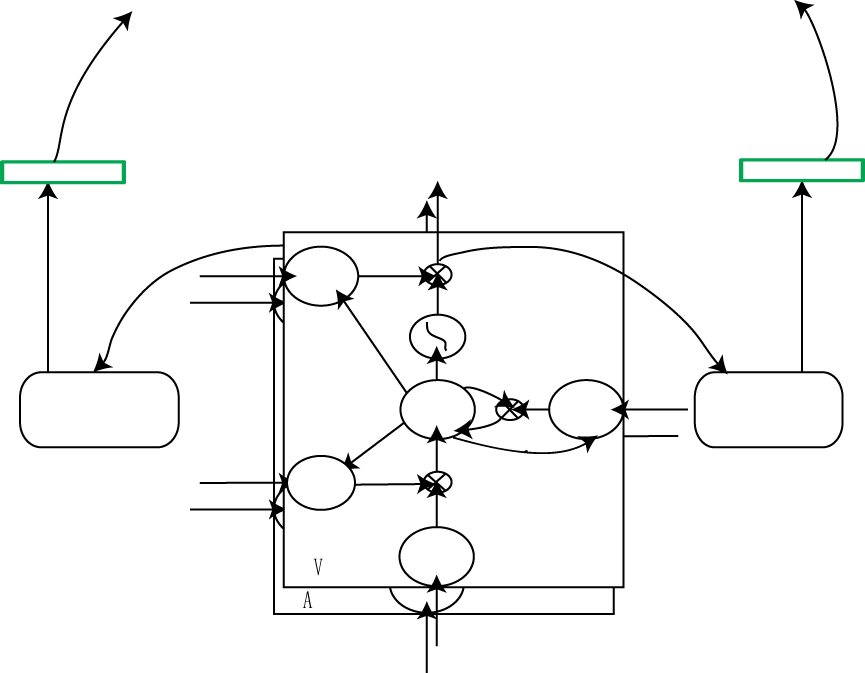}\\
\end{center}
\caption{The details of auxiliary connection.}
\end{figure}

\subsection{Training am-LSTM}

To train our am-LSTM model, we adopt the combination of squared multi-label margin loss.

\begin{equation}
loss(x,y)=\frac{1}{n}\sum^{n}_{i=1}max\{0,(1-x_i+y_i)^2\}
\end{equation}

Where $loss(x,y)$ is the squared multi-label margin loss, $x$ and $y$ are prediction and target.
\begin{equation}
\begin{split}
\label{equation:empirical}
Loss &= loss_{mian} + \alpha loss_{auxilary}^{video}+\beta loss_{auxilary}^{audio}\\
&= \frac{1}{n}\sum^{n}_{i=1}[max\{0,(1-x_{main}+y)^2\}+\alpha max\{0,(1-x_{aux}^{v}+y)^2\}+
                      \beta max\{0,(1-x_{aux}^{a}+y)^2\}]
\end{split}
\end{equation}

Where $Loss$ is the real training loss of am-LSTM, $loss_{main}$ is the loss of the main part, $loss_{auxliary}^{video}$ and $loss_{axuliary}^{audio}$ are the loss of auxiliary part. $\alpha$ and $\beta$ are the hyperparameters of the influence from auxiliary part. Here, we choose $\alpha = \beta =0.2$. And y is the classification target. The auxiliary connection is free enough so that the am-LSTM is flexible and extensive.
\section{Experiments}

In this section, we introduce our experiments details including datasets, data pre-processing, implementation details and some results, indicating that the am-LSTM is a robust, well-performing and flexible model for AVSR and likewise lipreading.

\subsection{Datasets}
We conducted our experiments in three datasets: AVLetters (Patterson et al., 2002), AVLetters2 (Cox et al., 2008) and AVDigits (Di Hu et al., 2015).

\begin{table}[htpb]
\centering
\begin{tabular}{|c|c|c|c|c|}
\hline
Datasets  & Speakers & Content & Times & Miscellaneous \\ \hline
AVLetters & 10       & A-Z     & 3     & pre-extracted \\ \hline
AVLetter2 & 5        & A-Z     & 7     & prvious split \\ \hline
AVDigits  & 6        & 0-9     & 9     & /             \\ \hline
\end{tabular}
\caption{Details of the datasets}
\label{my-label}
\end{table}
\subsection{Data Pre-processing}
If the video and audio are not splitted before, we split them into video and audio£¬and make video and audio the same length by truncation or completion. Eventually, centering the data.

\textbf{Video Pre-processing.} Firstly, the \emph{Viola-Jones algorithm}\cite{viola2001rapid} is employed to extract the Region-of-Interest surrounding the mouth. After the region is resized to $224\times224$ pixels, pre-trained VGG-16 \cite{simonyan2014very} is the tool to extract the image features. We use the features of last fully connected layer. Finally, reduce them to 100 principal components with PCA whitening.

\textbf{Audio Pre-processing.} The features of audio signal is extracted as spectrogram with 20ms Hamming window and 10ms overlap. With 251 points of Fast Fourier Transform and 50 principal components by PCA£¬the spectral coefficient vector is regraded as the audio features.

\textbf{Data Augmentation.} 4 contiguous audio frames correspond to 1 video frame in each time step. We let the aural and visual features in every video simultaneously shift 10 frames up and down randomly to double the data. As a result, the model will have better generalisation in time domain.

\subsection{Implementation Details}
The am-LSTM has two general LSTMs with dropout mapping 100 dimensional data to 50. Since the video and audio features have the same dimension, the projection is a identity matrix that could be omitted. The MLP in main phase has three layers with batch-normalization, together with ReLU activation function. The auxiliary phase has simple structure mapping 50 dimensional data to 10 classes prepared for classification. As mentioned, $\alpha = \beta =0.2$, the training loss is:

\begin{equation}
\begin{split}
\label{equation}
Loss &= \frac{1}{n}\sum^{n}_{i=1}[max\{0,(1-x_{main}+y)^2\}+0.2 max\{0,(1-x_{aux}^{v}+y)^2\}+
                      0.2 max\{0,(1-x_{aux}^{a}+y)^2\}]
\end{split}
\end{equation}

The am-LSTM is able to be trained and tested once. No redundant training process is needed. Lipreading is treated as cross modality speech recognition, that is, in the training phase, video and audio modalities are both needed, whereas in the testing phase, only video modality is presented.

\subsection{Results}
The Evaluation on am-LSTM is splitted into two parts: AVSR task and cross modality lipreading. AVSR is evaluated in two modalities and also trained in two modalities. However cross modality lipreading is evaluated in video but trained in video and audio both. Our experiments on AVSR is conducted in AVLetters2 and AVDigits, cross modality lipreading is conducted in AVLetters.
\subsubsection{Audio-visual Speech Recognition}
AVSR is the main task of our work, we conducted the experiments on AVLetters2 and AVDigits, the comparison methods are MDBN, MDAEs and RTMRBM. The results indicate that am-LSTM is much better than such models.
\begin{table}[htbp]

\begin{center}
\begin{tabular}{|l|c|c|}
\hline
Datasets                   & Model      & Mean Accuracy \\ \hline
\multirow{4}{*}{AVLetter2} & MDBN \cite{srivastava2012learning}       & 54.10\%       \\ \cline{2-3}
                           & MDAE \cite{ngiam2011multimodal}      & 67.89\%       \\ \cline{2-3}
                           & RTMRBM (Di Hu et al., 2015)    & 74.77\%         \\ \cline{2-3}
                           & am-LSTM & \textbf{89.11\%}       \\ \hline
\multirow{4}{*}{AVDigits}  & MDBN \cite{srivastava2012learning}      & 55.00\%       \\ \cline{2-3}
                           & MDAE \cite{ngiam2011multimodal}      & 66.74\%       \\ \cline{2-3}
                           & RTMRBM (Di Hu et al., 2015)    & 71.77\%       \\ \cline{2-3}
                           & am-LSTM & \textbf{85.23\%}       \\ \hline
\end{tabular}
\end{center}
\caption{AVSR performance on AVLetters2 and AVDigits. The result indicates that our model performs better than MDBN, MDAEs and RTMRBM. }
\label{label1}

\end{table}

\subsubsection{Cross Modality Lipreading}
Cross modality lipreading is the secondary task. As mentioned before, we trained am-LSTM in both modalities but evaluated in visual modality only. The experiments have two modes: only video and cross modality, which show the superiority of cross modality lipreading. In the experiments in cross modality lipreading, am-LSTM performs much better than MDAEs, CRBM, RTMRBM. The lipreading here is word-level lipreading.
\begin{table}[htbp]
\begin{center}
\begin{tabular}{|c|c|c|}
\hline
Mode           & Model                       & Mean  \\
                  & ~                           & Accuracy \\ \hline
\multirow{2}{*}{Only Video} & Multiscale Spatial Analysis \cite{matthews2002extraction}                       & 44.60\%       \\ \cline{2-3}
                   & Local Binary Pattern \cite{zhao2009lipreading}                     & 58.85\%       \\ \hline

\multirow{4}{*}{Cross Modality} & MDAEs \cite{ngiam2011multimodal}                       & 64.21\%       \\ \cline{2-3}
                   & CRBM \cite{amer2014multimodal}                       & 65.78\%       \\ \cline{2-3}
                   & RTMRBM (Di Hu et al., 2015)                       & 66.21\%       \\ \cline{2-3}
                   & am-LSTM                  & \textbf{88.83\%}       \\ \hline
\end{tabular}
\end{center}
\caption{Cross modality lipreading performance. The experiment suggest that cross modality lipreading is better than single modality. And result of our method performs much better than other model.  }
\label{label2}
\end{table}

\section{Conclusion}

We proposed an end-to-end deep model for AVSR and lipreading which increases mean accuracy obviously. Our experiments suggested that am-LSTM performs much better than other models in AVSR and cross modality lipreading. The benefits of am-LSTM are trained and tested once; extensibility and flexibility. There are no other training processes needed in training am-LSTM. Due to am-LSTM is simple, it could be a tool to model fusion information efficiently. Meanwhile, am-LSTM considers temporal connection, thus it is suitable for sequence features. In the future, we plan to apply am-LSTM in other multimodal temporal tasks and make it more flexible.

\subsubsection*{Acknowledgments}
I would like to acknowledge my friends Yi Sun, Wenqiang Yang and Peng Xie for helpful support. I would like to thank the developers of Torch\cite{collobert2011torch7} and Tensorflow\cite{abadi2016tensorflow}. I would also thank my laboratory for providing computational resources.

{\small
\bibliography{egbib}

\begin{thebibliography}{29}
\providecommand{\natexlab}[1]{#1}
\providecommand{\url}[1]{\texttt{#1}}
\expandafter\ifx\csname urlstyle\endcsname\relax
  \providecommand{\doi}[1]{doi: #1}\else
  \providecommand{\doi}{doi: \begingroup \urlstyle{rm}\Url}\fi

\bibitem[Abadi et~al.(2016)Abadi, Agarwal, Barham, Brevdo, Chen, Citro,
  Corrado, Davis, Dean, Devin, et~al.]{abadi2016tensorflow}
Mart{\i}n Abadi, Ashish Agarwal, Paul Barham, Eugene Brevdo, Zhifeng Chen,
  Craig Citro, Greg~S Corrado, Andy Davis, Jeffrey Dean, Matthieu Devin, et~al.
\newblock Tensorflow: Large-scale machine learning on heterogeneous distributed
  systems.
\newblock \emph{arXiv preprint arXiv:1603.04467}, 2016.

\bibitem[Amer et~al.(2014)Amer, Siddiquie, Khan, Divakaran, and
  Sawhney]{amer2014multimodal}
Mohamed~R Amer, Behjat Siddiquie, Saad Khan, Ajay Divakaran, and Harpreet
  Sawhney.
\newblock Multimodal fusion using dynamic hybrid models.
\newblock pages 556--563, 2014.

\bibitem[Amodei et~al.(2015)Amodei, Anubhai, Battenberg, Case, Casper,
  Catanzaro, Chen, Chrzanowski, Coates, Diamos, et~al.]{amodei2015deep}
Dario Amodei, Rishita Anubhai, Eric Battenberg, Carl Case, Jared Casper, Bryan
  Catanzaro, Jingdong Chen, Mike Chrzanowski, Adam Coates, Greg Diamos, et~al.
\newblock Deep speech 2: End-to-end speech recognition in english and mandarin.
\newblock \emph{arXiv preprint arXiv:1512.02595}, 2015.

\bibitem[Atrey et~al.(2010)Atrey, Hossain, El~Saddik, and
  Kankanhalli]{atrey2010multimodal}
Pradeep~K Atrey, M~Anwar Hossain, Abdulmotaleb El~Saddik, and Mohan~S
  Kankanhalli.
\newblock Multimodal fusion for multimedia analysis: a survey.
\newblock \emph{Multimedia systems}, 16\penalty0 (6):\penalty0 345--379, 2010.

\bibitem[Bengio et~al.(2015)Bengio, Goodfellow, and Courville]{bengio2015deep}
Yoshua Bengio, Ian~J Goodfellow, and Aaron Courville.
\newblock Deep learning.
\newblock \emph{An MIT Press book in preparation. Draft chapters available at
  http://www. iro. umontreal. ca/~ bengioy/dlbook}, 2015.

\bibitem[Collobert et~al.(2011)Collobert, Kavukcuoglu, and
  Farabet]{collobert2011torch7}
Ronan Collobert, Koray Kavukcuoglu, and Cl{\'e}ment Farabet.
\newblock Torch7: A matlab-like environment for machine learning.
\newblock In \emph{BigLearn, NIPS Workshop}, number EPFL-CONF-192376, 2011.

\bibitem[Graves et~al.(2006)Graves, Fern{\'a}ndez, Gomez, and
  Schmidhuber]{graves2006connectionist}
Alex Graves, Santiago Fern{\'a}ndez, Faustino Gomez, and J{\"u}rgen
  Schmidhuber.
\newblock Connectionist temporal classification: labelling unsegmented sequence
  data with recurrent neural networks.
\newblock In \emph{Proceedings of the 23rd international conference on Machine
  learning}, pages 369--376. ACM, 2006.

\bibitem[He et~al.(2015)He, Zhang, Ren, and Sun]{he2015deep}
Kaiming He, Xiangyu Zhang, Shaoqing Ren, and Jian Sun.
\newblock Deep residual learning for image recognition.
\newblock \emph{arXiv preprint arXiv:1512.03385}, 2015.

\bibitem[Hochreiter and Schmidhuber(1997)]{hochreiter1997long}
Sepp Hochreiter and J{\"u}rgen Schmidhuber.
\newblock Long short-term memory.
\newblock \emph{Neural computation}, 9\penalty0 (8):\penalty0 1735--1780, 1997.

\bibitem[Krizhevsky et~al.(2012)Krizhevsky, Sutskever, and
  Hinton]{krizhevsky2012imagenet}
Alex Krizhevsky, Ilya Sutskever, and Geoffrey~E Hinton.
\newblock Imagenet classification with deep convolutional neural networks.
\newblock In \emph{Advances in neural information processing systems}, pages
  1097--1105, 2012.

\bibitem[Mao et~al.(2014)Mao, Xu, Yang, Wang, Huang, and Yuille]{mao2014deep}
Junhua Mao, Wei Xu, Yi~Yang, Jiang Wang, Zhiheng Huang, and Alan Yuille.
\newblock Deep captioning with multimodal recurrent neural networks (m-rnn).
\newblock \emph{arXiv preprint arXiv:1412.6632}, 2014.

\bibitem[Maragos et~al.(2008)Maragos, Potamianos, and
  Gros]{maragos2008multimodal}
Petros Maragos, Alexandros Potamianos, and Patrick Gros.
\newblock \emph{Multimodal processing and interaction: audio, video, text},
  volume~33.
\newblock Springer Science \& Business Media, 2008.

\bibitem[Matthews et~al.(2002)Matthews, Cootes, Bangham, Cox, and
  Harvey]{matthews2002extraction}
Iain Matthews, Timothy~F Cootes, J~Andrew Bangham, Stephen Cox, and Richard
  Harvey.
\newblock Extraction of visual features for lipreading.
\newblock \emph{IEEE Transactions on Pattern Analysis and Machine
  Intelligence}, 24\penalty0 (2):\penalty0 198--213, 2002.

\bibitem[Mroueh et~al.(2015)Mroueh, Marcheret, and Goel]{mroueh2015deep}
Youssef Mroueh, Etienne Marcheret, and Vaibhava Goel.
\newblock Deep multimodal learning for audio-visual speech recognition.
\newblock pages 2130--2134, 2015.

\bibitem[Nath and Beauchamp(2012)]{nath2012a}
Audrey~R Nath and Michael~S Beauchamp.
\newblock A neural basis for interindividual differences in the mcgurk effect,
  a multisensory speech illusion.
\newblock \emph{NeuroImage}, 59\penalty0 (1):\penalty0 781--787, 2012.

\bibitem[Nefian et~al.(2002)Nefian, Liang, Pi, Liu, and
  Murphy]{nefian2002dynamic}
Ara~V Nefian, Luhong Liang, Xiaobo Pi, Xiaoxing Liu, and Kevin Murphy.
\newblock Dynamic bayesian networks for audio-visual speech recognition.
\newblock \emph{EURASIP Journal on Advances in Signal Processing},
  2002\penalty0 (11):\penalty0 1--15, 2002.

\bibitem[Ngiam et~al.(2011)Ngiam, Khosla, Kim, Nam, Lee, and
  Ng]{ngiam2011multimodal}
Jiquan Ngiam, Aditya Khosla, Mingyu Kim, Juhan Nam, Honglak Lee, and Andrew~Y
  Ng.
\newblock Multimodal deep learning.
\newblock In \emph{Proceedings of the 28th international conference on machine
  learning (ICML-11)}, pages 689--696, 2011.

\bibitem[Noda et~al.(2015)Noda, Yamaguchi, Nakadai, Okuno, and
  Ogata]{noda2015audio-visual}
Kuniaki Noda, Yuki Yamaguchi, Kazuhiro Nakadai, Hiroshi~G Okuno, and Tetsuya
  Ogata.
\newblock Audio-visual speech recognition using deep learning.
\newblock \emph{Applied Intelligence}, 42\penalty0 (4):\penalty0 722--737,
  2015.

\bibitem[Pascanu et~al.(2013)Pascanu, Mikolov, and
  Bengio]{pascanu2013difficulty}
Razvan Pascanu, Tomas Mikolov, and Yoshua Bengio.
\newblock On the difficulty of training recurrent neural networks.
\newblock \emph{ICML (3)}, 28:\penalty0 1310--1318, 2013.

\bibitem[Prechelt(1998)]{prechelt1998automatic}
Lutz Prechelt.
\newblock Automatic early stopping using cross validation: quantifying the
  criteria.
\newblock \emph{Neural Networks}, 11\penalty0 (4):\penalty0 761--767, 1998.

\bibitem[Rabiner(1989)]{rabiner1989tutorial}
Lawrence~R Rabiner.
\newblock A tutorial on hidden markov models and selected applications in
  speech recognition.
\newblock \emph{Proceedings of the IEEE}, 77\penalty0 (2):\penalty0 257--286,
  1989.

\bibitem[Simonyan and Zisserman(2014)]{simonyan2014very}
Karen Simonyan and Andrew Zisserman.
\newblock Very deep convolutional networks for large-scale image recognition.
\newblock \emph{arXiv preprint arXiv:1409.1556}, 2014.

\bibitem[Srivastava and
  Salakhutdinov(2012{\natexlab{a}})]{srivastava2012learning}
Nitish Srivastava and Ruslan Salakhutdinov.
\newblock Learning representations for multimodal data with deep belief nets.
\newblock In \emph{International conference on machine learning workshop},
  2012{\natexlab{a}}.

\bibitem[Srivastava and
  Salakhutdinov(2012{\natexlab{b}})]{srivastava2012multimodal}
Nitish Srivastava and Ruslan~R Salakhutdinov.
\newblock Multimodal learning with deep boltzmann machines.
\newblock In \emph{Advances in neural information processing systems}, pages
  2222--2230, 2012{\natexlab{b}}.

\bibitem[Srivastava et~al.(2014)Srivastava, Hinton, Krizhevsky, Sutskever, and
  Salakhutdinov]{srivastava2014dropout}
Nitish Srivastava, Geoffrey~E Hinton, Alex Krizhevsky, Ilya Sutskever, and
  Ruslan Salakhutdinov.
\newblock Dropout: a simple way to prevent neural networks from overfitting.
\newblock \emph{Journal of Machine Learning Research}, 15\penalty0
  (1):\penalty0 1929--1958, 2014.

\bibitem[Sutskever et~al.(2014)Sutskever, Vinyals, and
  Le]{sutskever2014sequence}
Ilya Sutskever, Oriol Vinyals, and Quoc~V Le.
\newblock Sequence to sequence learning with neural networks.
\newblock In \emph{Advances in neural information processing systems}, pages
  3104--3112, 2014.

\bibitem[Tamura et~al.(2015)Tamura, Ninomiya, Kitaoka, Osuga, Iribe, Takeda,
  and Hayamizu]{tamura2015audio}
Satoshi Tamura, Hiroshi Ninomiya, Norihide Kitaoka, Shin Osuga, Yurie Iribe,
  Kazuya Takeda, and Satoru Hayamizu.
\newblock Audio-visual speech recognition using deep bottleneck features and
  high-performance lipreading.
\newblock In \emph{2015 Asia-Pacific Signal and Information Processing
  Association Annual Summit and Conference (APSIPA)}, pages 575--582. IEEE,
  2015.

\bibitem[Viola and Jones(2001)]{viola2001rapid}
Paul Viola and Michael Jones.
\newblock Rapid object detection using a boosted cascade of simple features.
\newblock In \emph{Computer Vision and Pattern Recognition, 2001. CVPR 2001.
  Proceedings of the 2001 IEEE Computer Society Conference on}, volume~1, pages
  I--511. IEEE, 2001.

\bibitem[Zhao et~al.(2009)Zhao, Barnard, and Pietikainen]{zhao2009lipreading}
Guoying Zhao, Mark Barnard, and Matti Pietikainen.
\newblock Lipreading with local spatiotemporal descriptors.
\newblock \emph{IEEE Transactions on Multimedia}, 11\penalty0 (7):\penalty0
  1254--1265, 2009.

\end{thebibliography}
\bibliographystyle{plainnat}}

\end{document}